\documentclass[conference]{IEEEtran}
\IEEEoverridecommandlockouts
\usepackage{cite}
\usepackage{amsmath,amssymb,amsfonts}
\usepackage{graphicx}
\usepackage{textcomp}
\usepackage{xcolor}
\usepackage{booktabs}
\usepackage{array}
\usepackage{url}
\usepackage{algpseudocode}
\algrenewcommand\algorithmicrequire{\textbf{Input:}}
\algrenewcommand\algorithmicensure{\textbf{Output:}}
\algrenewcommand\alglinenumber[1]{\scriptsize #1:}
\graphicspath{{figures/}}
\newcommand{\method}{OGR-MARL}
\DeclareMathOperator{\sat}{sat}
\newcounter{algorithm}

\def\BibTeX{{\rm B\kern-.05em{\sc i\kern-.025em b}\kern-.08em
    T\kern-.1667em\lower.7ex\hbox{E}\kern-.125emX}}

\makeatletter
\g@addto@macro\normalsize{%
  \setlength{\abovedisplayskip}{3pt}
  \setlength{\belowdisplayskip}{3pt}
  \setlength{\abovedisplayshortskip}{2pt}
  \setlength{\belowdisplayshortskip}{2pt}
}
\makeatother

\begin{document}

\title{\method{}: Option-Guided Residual Multi-Agent Reinforcement Learning for Heterogeneous USV Cooperative Pursuit in Constrained Port Waterways}

\author{\IEEEauthorblockN{Jiayang Mao{$\dagger$}{*}}
\IEEEauthorblockA{\textit{College of Information Engineering} \\
\textit{Sichuan Agricultural University}\\
Yaan, China \\
maojiayang2024@163.com}
\and
\IEEEauthorblockN{Lanfeng Wang{$\dagger$}}
\IEEEauthorblockA{\textit{Shenzhen International Graduate School} \\
\textit{Tsinghua University}\\
Shenzhen, China \\
wanglf24@mails.tsinghua.edu.cn}
\and
\IEEEauthorblockN{Zhao-Han Peng}
\IEEEauthorblockA{\textit{Shenzhen International Graduate School} \\
\textit{Tsinghua University}\\
Shenzhen, China \\
pzh23@mails.tsinghua.edu.cn}

\thanks{{$\dagger$} {Equal contribution}}
\thanks{{*} {Corresponding author}}
}

\maketitle
\raggedbottom

\begin{abstract}
Heterogeneous USV cooperative pursuit in constrained port waterways requires evader interception under navigation, traffic, and role constraints. This paper proposes \method{}, an option-guided residual multi-agent reinforcement learning framework that is decoupled from a specific MARL algorithm. \method{} integrates shared evader belief, role-conditioned option targets, adaptive rule penalties, and residual policy learning, allowing different MARL algorithms to learn corrective actions on top of rule-guided behaviors rather than exploring constrained port environments from scratch. We instantiate \method{} with representative continuous-control MARL backbones, including MADDPG, MATD3, MAPPO, and MASAC, yielding OGR-MADDPG, OGR-MATD3, OGR-MAPPO, and OGR-MASAC. Experiments in an abstract Xiazhimen port-waterway scenario show that the OGR-MASAC instantiation achieves a 75.0\% capture rate, promising mission-effective rule compliance, and the best heterogeneous coordination among the tested methods. Without retraining, zero-shot transfer to a QGIS/AIS-informed Xiazhimen map achieves promising results, demonstrating the generalization potential of \method{} in more complex port scenarios.
\end{abstract}

\begin{IEEEkeywords}
cooperative pursuit, multi-agent reinforcement learning, option guidance, residual learning, heterogeneous USV
\end{IEEEkeywords}

\section{Introduction}
Unmanned surface vehicles (USVs) are increasingly vital for maritime surveillance, environmental monitoring, and harbor security because they can operate persistently in waters where human operation is costly or risky \cite{b1}. Multi-USV coordination has received sustained attention as a way to improve robustness and efficiency in complex marine missions, among which the cooperative pursuit problem constitutes a significant class of issues \cite{b15}. However, pursuit in constrained port-waterways is more demanding than in open environments. Pursuers must navigate narrow lanes, avoid shores, anchorages, and moving traffic while capturing an evader.

Existing pursuit studies often focus on simplified or open scenarios. Classical model-based methods, such as artificial potential fields (APF), model predictive control, and path planning, provide interpretable guidance and COLREGs-aware collision avoidance \cite{b3,b4,b22,b23}. Yet, they require scenario-specific tuning and struggle with partial observability, role heterogeneity, and online role switching.

Conversely, multi-agent reinforcement learning (MARL) offers a data-driven approach for environment exploration and cooperative pursuit \cite{b5,b9,b10}. Despite some successful applications in multi-USV cooperative pursuit \cite{b16,b17,b25}, directly applying MARL to constrained port scenarios from scratch remains challenging. A pure policy-gradient learner may suffer from sparse rewards, frequent rule violations, and exploration inefficiency, often expending budgets on collisions or failing to leave the harbor. Conversely, a pure rule-based controller is interpretable but lacks adaptability against tactical evader replanning.

To bridge this gap, we propose option-guided multi-agent reinforcement learning (\method{}), an algorithm-agnostic framework for heterogeneous multi-USV cooperative pursuit in constrained port-waterways. The pursuing team is heterogeneous: interceptors have capture capability with higher speed, while scouts have a larger sensing radius for target visibility. A* search \cite{b11} and APF are utilized to model evader behavior and generate high-level option targets. Motivated by the option framework for temporal decomposition \cite{b6} and residual reinforcement learning with rule-guided policies \cite{b13}, our approach integrates an option guidance module, which provides high-level modes and geometric targets, with a general MARL algorithm that learns residual continuous actions and coordination patterns for the heterogeneous pursuing team. This design effectively corrects timing, local motion, and interaction effects beyond rule-based guidance while preserving structural safety priors and learned adaptability \cite{b18}.

The main contributions are as follows:
\begin{itemize}
    \item We establish a benchmark scenario for heterogeneous multi-USV cooperative pursuit in constrained port waterways, incorporating shores, anchorage areas, bidirectional traffic lanes, buffer regions, and moving cargo vessels.
    \item We propose \method{}, an algorithm-agnostic option-guided residual MARL framework that integrates shared evader belief, role-conditioned option targets, and adaptive rule penalties, and can be instantiated with different MARL backbones.
    \item We evaluate the proposed framework through backbone comparison, reward ablation, and zero-shot transfer to a QGIS/AIS-informed Xiazhimen map.
\end{itemize}

\section{Problem Formulation}
\subsection{Simulated Port Waterway Scenario}
\begin{figure}[t]
\centering
\includegraphics[width=\linewidth]{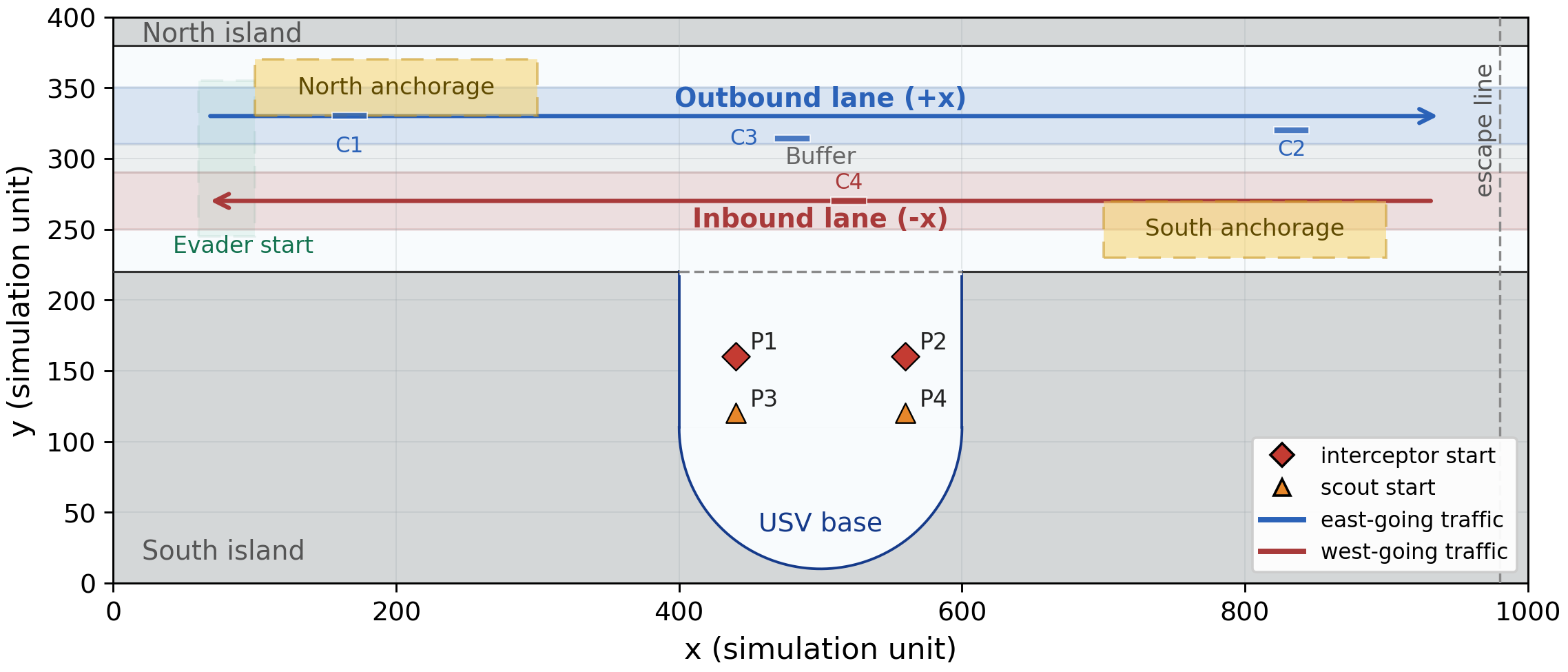}
\caption{Abstracted Xiazhimen port waterways.}
\label{fig:scenario}
\end{figure}

The simulation scenario is an abstracted Xiazhimen port waterway (in Zhejiang Province of China) on a $1000\times400$ domain. As shown in Fig.~\ref{fig:scenario}, the channel is bounded by the north and south islands, with a USV base located at $x\in[400,600]$. Two anchorage areas occupy $[100,300]\times[330,370]$ and $[700,900]\times[230,270]$. The outbound and inbound traffic lanes occupy $y\in[310,350]$ and $[250,290]$, respectively, and a buffer area occupies $y\in[290,310]$. The line $x=980$ serves as the eastern escape boundary, and an episode terminates as an escape when the evader crosses this boundary.

The pursuer team is denoted by $\mathcal{N}=\{P_1,P_2,P_3,P_4\}$. Pursuers $P_1$ and $P_2$ are fast interceptors and the only USVs with capture capability, while $P_3$ and $P_4$ are slower scouts with larger sensing radius. Four cargo vessels move along the center lines of the traffic lanes:
\begin{equation}
x^c_{k,t+1}=\left(x^c_{k,t}+v^c_k\Delta t\right)\bmod L_x,\quad y^c_{k,t+1}=y^c_{k,t},
\end{equation}
where $L_x=1000$. The three outbound vessels move eastward at a speed of 4.69 simulation units per second, while the inbound vessel moves westward at a speed of -6.26.

\subsection{Pursuer Dynamics and Actions}
Each pursuer is modeled by a three-degree-of-freedom (3-DOF) motion model \cite{b12}. Let
\begin{equation}
    {\boldsymbol \eta}_i=[x_i,y_i,\psi_i]^T,\quad
    {\boldsymbol\nu}_i=[u_i,v_i,r_i]^T,
\end{equation}
where $(x_i,y_i)$ are the position coordinates, $\psi_i$ is the heading angle, and $(u_i,v_i,r_i)$ denote the surge velocity, sway velocity, and yaw rate, respectively. The dynamic equations are
\begin{equation}
\dot{{\boldsymbol \eta}}_i = {\boldsymbol J}(\psi_i){\boldsymbol \nu}_i,\quad {\boldsymbol M}_i\dot{{\boldsymbol \nu}}_i + {\boldsymbol C}_i{\boldsymbol \nu}_i + {\boldsymbol D}_i{\boldsymbol \nu}_i = {\boldsymbol \tau}_i,
\end{equation}
\begin{equation}
{\boldsymbol J}(\psi_i)=
\begin{bmatrix}
\cos\psi_i & -\sin\psi_i & 0\\
\sin\psi_i & \cos\psi_i & 0\\
0 & 0 & 1
\end{bmatrix}.
\end{equation}
Here, ${\boldsymbol M}_i$, ${\boldsymbol C}_i$, and ${\boldsymbol D}_i$ denote the inertia, Coriolis--centripetal, and nonlinear damping terms, respectively. 

\begin{table}[t]
\caption{Parameters of the Pursuers}
\begin{center}
\begin{tabular}{ccc}
\toprule
\textbf{Parameter} & \textbf{Interceptor} & \textbf{Scout} \\
\midrule
Normalized mass $m$ & 380 & 760 \\
Normalized yaw inertia $I_z$ & 620 & 1760 \\
Maximum surge speed $u^{\max}$ (sim. units/s) & 10.0 & 4.6 \\
Maximum heading increment (rad) & 0.580 & 0.260 \\
Maximum thrust & 1320 & 520 \\
Sensing radius (sim. units) & 8 & 180 \\
Capture capability & Yes & No \\
\bottomrule
\end{tabular}
\label{tab:role_params}
\end{center}
\end{table}

A fourth-order Runge--Kutta integrator is used for the inner-loop dynamics. A low-level speed-heading controller maps each high-level MARL action of the pursuers to a surge force and yaw moment. The high-level action of pursuer $i$ is
\begin{equation}
{\boldsymbol a}_i
=[a_i^v,a_i^\psi]^T,
\end{equation}
where $a_i^v$ is related to the surge speed and $a_i^\psi$ is related to the heading angle. The desired surge speed and heading angle are
\begin{equation}
u_{d,i}=\frac{u^{\max}_i}{2}(a_i^v+1),\quad \psi_{d,i}=\psi_i+a_i^\psi\Delta\psi^{\max}_i.
\end{equation}
Let $e^u_i=u_{d,i}-u_i$ and $e^\psi_i=\operatorname{wrap}(\psi_{d,i}-\psi_i)$, where the wrap function normalizes the heading-angle difference to the principal interval, \emph{i.e.}, $(-\pi,\pi]$. The controller computes
\begin{align}
\tau_{x,i} &= \sat\left(K^u_{p,i}e^u_i+K^u_{I,i}\int e^u_i dt,\;0,\;2T_i^{\max}\right),\\
\tau_{z,i} &= \sat\left(K^\psi_{p,i}e^\psi_i-K^\psi_{d,i}r_i,\;-\tau_{z,i}^{\max},\;\tau_{z,i}^{\max}\right),
\end{align}
where $\sat(\cdot,l,u)$ denotes saturation to $[l,u]$. The corresponding port and starboard thrust commands are
\begin{align}
T_{p,i}=\sat\left(\frac{\tau_{x,i}}{2}+\frac{\tau_{z,i}}{B_i},0,T_i^{\max}\right),\\
T_{s,i}=\sat\left(\frac{\tau_{x,i}}{2}-\frac{\tau_{z,i}}{B_i},0,T_i^{\max}\right).
\end{align}
The resulting input force is
\begin{equation}
{\boldsymbol \tau}_i=[T_{p,i}+T_{s,i}, 0, (T_{p,i}-T_{s,i})B_i/2]^T.
\end{equation}

The parameters of the pursuers are listed in Table~\ref{tab:role_params}. Scouts provide visibility in a wide area and maintain the shared evader belief, whereas interceptors rely on close-range perception and the team-level belief for final interception. The heterogeneous system encourages scouts to detect the evader early.

\subsection{The Evader}
The evader is modeled as a mass point with constrained acceleration and yaw rate. Its state is ${\boldsymbol s}_e=[x_e,y_e,\psi_e,v_e]^T$. Given the desired speed $v_d$ and heading angle $\psi_d$, define
\begin{align}
    \Delta v_t &= \sat(v_d-v_{e,t},-a^{\max}_e\Delta t,a^{\max}_e\Delta t),\\
    \Delta \psi_t &= \sat(\operatorname{wrap}(\psi_d-\psi_{e,t}),-\omega^{\max}_e\Delta t,\omega^{\max}_e\Delta t).
\end{align}
The evader's state is updated by
\begin{align}
    v_{e,t+1} &= \sat(v_{e,t}+\Delta v_t,0,v^{\max}_e),\\
    \psi_{e,t+1} &= \operatorname{wrap}(\psi_{e,t}+\Delta\psi_t),\\
    x_{e,t+1} &= x_{e,t}+v_{e,t+1}\cos\psi_{e,t+1}\Delta t,\\
    y_{e,t+1} &= y_{e,t}+v_{e,t+1}\sin\psi_{e,t+1}\Delta t.
\end{align}

The evader uses an A*-based global planner and an APF-based local collision-avoidance module. The A* layer plans a path toward the eastern escape boundary while taking into account the shores, cargo vessels, and inflated safety regions around pursuers. The APF layer combines attraction to the current waypoint planned by A* with repulsion from shores, cargo vessels, and nearby pursuers. The desired heading angle is the same as the direction of the force generalized by APF:
\begin{equation}
F_e=F_{\mathrm{goal}}+\sum_{h\in\mathcal{H}}F_{\mathrm{rep}}^{h}+\sum_{i\in\mathcal{N}}F_{\mathrm{rep}}^{i},
\end{equation}
where $\mathcal{H}$ contains the shore boundary and cargo vessels, and the desired heading angle is obtained from the force direction.

\subsection{Termination Condition}
An episode ends with capture, escape, or timeout. Capture is declared if any interceptor satisfies:
\begin{equation}
\min_{i\in\{P_1,P_2\}}\left\|(x_i,y_i)-(x_e,y_e)\right\|_2 \leq R_c,
\end{equation}
where $R_c$ is a distance threshold. Scouts may detect and track the evader, but they do not have capture capability. 

The evader escapes when it crosses the eastern escape boundary. Timeout happens when the maximum episode length is reached without capture or escape. 

\subsection{Partial Observation}
The cooperative pursuit task for the pursuers is partially observable because only scouts have long-range sensing capability: Interceptors have a sensing radius of 8 simulation units, whereas scouts have a sensing radius of 180. The team maintains a shared evader belief from the latest evader observation. When any USV detects the evader, the belief is updated. Otherwise, the belief is propagated by dead reckoning using the last known speed and heading angle:
\begin{equation}
    \hat{x}_{t+1}=\hat{x}_{t}+\hat{v}_{t}\cos \hat{\psi}_{t}\Delta t,\quad
    \hat{y}_{t+1}=\hat{y}_{t}+\hat{v}_{t}\sin \hat{\psi}_{t}\Delta t.
\end{equation}

The belief confidence $c_t$ decays exponentially with the number of steps since the last observation:
\begin{equation}
c_t=\exp(-n_t/\tau_b),
\end{equation}
where $n_t$ is the number of steps since the last observation and $\tau_b$ is a decay constant. 

Each pursuer receives a 61-dimensional observation vector consisting of 7 self-state variables, 12 relative teammate variables, 7 evader belief variables, 20 cargo vessel variables, 5 local geometry and rule compliance variables, a 5-dimensional option one-hot vector, 4 option-guidance variables
comprising the relative option-target position, the target distance,
and the distance to the shared evader belief, and the normalized
episode time. 

\section{\method{} Algorithm}
\subsection{CTDE Framework}
\begin{figure}[t]
\centering
\includegraphics[width=\linewidth]{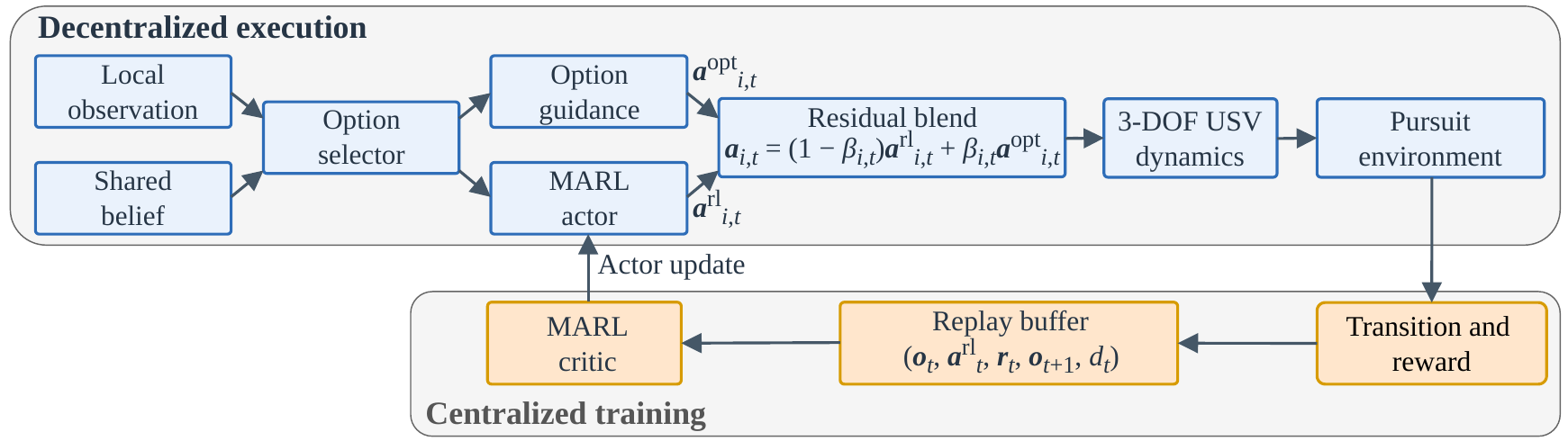}
\caption{Information flow in \method{}.}
\label{fig:framework}
\end{figure}

The cooperative pursuit task is formulated as a partially observable multi-agent Markov decision process. We adopt the centralized training and decentralized execution (CTDE) paradigm for \method{}, as illustrated in Fig.~\ref{fig:framework}. During training, a centralized critic utilizes global information, while each actor executes using only local observations during decentralized execution. Each actor outputs a two-dimensional continuous residual control action. Due to heterogeneous dynamics, sensing ranges, and functional roles, we adopt independent actor--critic networks without parameter sharing.

In the general \method{} formulation, each pursuer policy is defined as
\begin{equation}
\pi_i(\boldsymbol{a}_i^{\mathrm{rl}} \mid \boldsymbol{o}_i),
\end{equation}
where $\boldsymbol{a}_i^{\mathrm{rl}}$ denotes the residual action produced by a generic MARL backbone conditioned on local observation $\boldsymbol{o}_i$.

$\Omega_i$, $G_i$, and $\Gamma_i$ denote the option selector,
option-action generator, and blending rule of pursuer $i$, respectively. These modules are independent of the MARL algorithm and remain unchanged across different instantiations of \method{}.

%Algorithm 1 begins
\begin{center}
\begin{minipage}{0.98\linewidth}
\hrule height 0.8pt
\vspace{2pt}
\refstepcounter{algorithm}
\noindent\textbf{Algorithm \thealgorithm} \method{} Training and Evaluation
\label{alg:ogmarL}
\vspace{2pt}
\hrule height 0.4pt
\vspace{2pt}

\footnotesize
\begin{algorithmic}[1]
\Statex \textbf{Input:} $\mathcal{E}$, $\mathcal{N}$, $\mathcal{S}$,
$\mathcal{O}$, and $\mathcal{K}_{\mathrm{eval}}$
\Statex \textbf{Output:}
$\Pi^{\mathrm{OG}}=\{\pi_i,\Omega_i,G_i,\Gamma_i\}_{i\in\mathcal{N}}$

\State Initialize the MARL algorithm parameters $\pi_i$, $Q_i$, and (if applicable) auxiliary parameters
\State Initialize replay buffer $\mathcal{D}$ and curriculum variables

\Statex \textbf{Training:}
\For{each training episode}
    \State Select scenario stage $S$ according to curriculum schedule
    \State Reset environment $\mathcal{E}_S$, initialize belief $b_0$, set $t \gets 0$
    \Repeat
        \State Update shared evader belief $b_t$ and confidence $c_t$

        \State Select option $m_{i,t}$ via $\Omega_i$ and generate option action $\boldsymbol{a}_{i,t}^{\mathrm{opt}}$ via $G_i$

        \State Construct joint observation $\boldsymbol{o}_t = (\boldsymbol{o}_{i,t})_{i\in\mathcal{N}}$

        \State Sample residual action $\boldsymbol{a}_{i,t}^{\mathrm{rl}} \sim \pi_i(\cdot \mid \boldsymbol{o}_{i,t})$

        \State Blend actions via $\Gamma_i$ to obtain executed action $\boldsymbol{a}_{i,t}$

        \State Execute $\boldsymbol{a}_t$, observe $\boldsymbol{o}_{t+1}$ and terminal flag $d_t$

        \State Update curriculum statistics and rule violation statistics if needed

        \State Compute individual reward $r^{\mathrm{ind}}_{i,t}$ and mixed reward $r_t$

        \State Store transition $(\boldsymbol{o}_t,\boldsymbol{a}_t^{\mathrm{rl}},\boldsymbol{r}_t,\boldsymbol{o}_{t+1},d_t)$ in $\mathcal{D}$

        \State Update the MARL parameters according to its optimization rule

        \State $t \gets t + 1$
    \Until{$d_t = 1$ (capture, escape, or timeout)}
\EndFor

\Statex \textbf{Evaluation:}
\State Evaluate $\Pi^{\mathrm{OG}}$ in $\mathcal{E}_{S_4}$ over $\mathcal{K}_{\mathrm{eval}}$ without parameter updates
\State Report SR, TC, MRC, and HCS
\end{algorithmic}

\vspace{2pt}
\hrule height 0.8pt
\end{minipage}
\end{center}
%Algorithm 1 ends

\subsection{Option Guidance}
Exploration over the aforementioned continuous action space is inefficient in constrained port waterway environments. To address this issue, we introduce a lightweight option layer on top of a generic MARL backbone. The option set is
\begin{equation}
\mathcal{O}=\{\text{search},\text{track},\text{intercept},\text{block},\text{recover}\}.
\end{equation}

The option layer assigns each pursuer an execution mode according to its role, rule-compliance state, evader visibility, and belief confidence. Scouts select the search or track option to maintain channel coverage and update the shared evader belief. Interceptors select the intercept or block option to form a pincer maneuver, close in on the evader, and complete the capture. The recover option guides a pursuer away from shores, anchorage areas, and buffer regions, or prevents violations of traffic lane direction constraints.

Within the proposed \method{} framework, the option module provides a structured action prior that is independent of the underlying MARL algorithm. Let ${\boldsymbol a}_i^{\mathrm{rl}}$ denote the residual action produced by a generic MARL policy conditioned on local observations, and let ${\boldsymbol a}_i^{\mathrm{opt}}$ denote the option-guided action. The final executed action is obtained via a bounded blending mechanism:
\begin{equation}
{\boldsymbol a}_i
=\operatorname{clip}\left(
(1-\beta_i){\boldsymbol a}_i^{\mathrm{rl}}
+\beta_i{\boldsymbol a}_i^{\mathrm{opt}},-1,1\right).
\end{equation}

\subsection{Structured Reward and Curriculum Learning}

To train \method{} in constrained port waterways, we use a backbone-agnostic training design that combines structured reward shaping with curriculum learning. The reward provides task-, rule-, and role-level feedback, whereas the curriculum gradually increases task difficulty and constraint strength.

The individual reward of pursuer $i$ is decomposed into four components:
\begin{equation}
r^{\rm ind}_{i,t}
=
\chi_t\left(
r^{\rm T}_{i,t}
+
r^{\rm R}_{i,t}
+
r^{\rm H}_{i,t}
\right)
+
(1-\chi_t)r^{\rm term}_{i,t},
\end{equation}
where $\chi_t\in\{0,1\}$ is a non-terminal indicator, \textit{i.e.}, $\chi_t=1$ before the termination and $\chi_t=0$ at the terminal transition. The task-progress reward $r^{\rm T}_{i,t}$ encourages pursuit behaviors that are essential for completing the mission, including harbor departure, evader tracking, interception, and pincer formation. The rule-compliance term $r^{\rm R}_{i,t}$ penalizes violations of port-waterway constraints, such as shore intrusion and anchorage intrusion. The heterogeneous-role reward $r^{\rm H}_{i,t}$ promotes complementary behaviors between scouts and interceptors, encouraging scouts to maintain evader visibility and interceptors to contribute to distance closing and capture. The terminal reward $r^{\rm term}_{i,t}$ evaluates the final mission outcome, including successful capture, evader escape, and timeout. 

To promote cooperative behavior, the training reward blends individual feedback with team-level feedback:

\begin{equation}
r_{i,t}=\lambda r^{\rm ind}_{i,t}+(1-\lambda)\frac{1}{|\mathcal{N}|}\sum_{j\in \mathcal{N}}r^{\rm ind}_{j,t}.
\end{equation}

\begin{table}[t]
\caption{Curriculum Learning Stages}
\begin{center}
\resizebox{\linewidth}{!}{
\begin{tabular}{cccccc}
\toprule
\textbf{Stage} & \textbf{Start Step} & \textbf{Evader Speed} & \textbf{Repulsion Scale} & \textbf{Capture Radius} & \textbf{Penalty Scale} \\
\midrule
S0 & 0 & 4.75 & 0.25 & 8.00 & 0.15 \\
S1 & 16k & 5.00 & 0.45 & 6.50 & 0.35 \\
S2 & 56k & 5.10 & 0.62 & 5.90 & 0.60 \\
S3 & 288k & 5.10 & 0.85 & 5.45 & 0.82 \\
S4 & 400k & 5.10 & 1.00 & 5.00 & 1.00 \\
\bottomrule
\end{tabular}
}
\label{tab:curriculum}
\end{center}
\end{table}

Directly training on the hardest pursuit setting is inefficient because the evader moves actively, only interceptors have capture capability, and the capture radius is small. Therefore, we adopt a five-stage curriculum learning strategy \cite{b20}, as summarized in Table~\ref{tab:curriculum}. Across stages, the evader becomes more difficult to capture, the capture condition becomes stricter, and the rule-compliance penalty is gradually strengthened. This schedule first allows the agents to discover basic harbor-exit, tracking, and pursuit behaviors under relatively permissive conditions, and then shifts learning toward precise interception and rule-compliant heterogeneous coordination.

\section{Experiments}
\subsection{Experimental Setup}
The algorithms are trained using eight parallel environments. The parameters for MARL algorithms are illustrated in Table~\ref{tab:marl_params}. MARL baselines execute only learned actions without option-guided action blending; option-related observation entries are kept only to preserve the same input schema. Therefore, the comparison isolates the effect of option-action execution rather than removing all option-related auxiliary inputs. 
\begin{table}[t]
\caption{Parameters for MARL Algorithms}
\begin{center}
\begin{tabular}{cc|cc}
\toprule
\textbf{Parameter} & \textbf{Value} &\textbf{Parameter} & \textbf{Value} \\
\midrule
Actor hidden sizes & $(256, 256)$ & Learning rates & $3\times 10^{-4}$ \\
Critic hidden sizes & $(512, 512)$ & Time step & $0.5\,\mathrm{s}$  \\
Target network update rate & $5\times 10^{-3}$ & Replay buffer & $1\times 10^{6}$ \\
Reward discount factor & $0.995$ & Batch size & $512$ \\
Episode step & $8\times 10^{3}$ & Total step & $7\times 10^{5}$ \\
\bottomrule
\end{tabular}
\label{tab:marl_params}
\end{center}
\end{table}

\subsection{Evaluation Metrics}
The evaluation takes 100 test episodes with 100 seeds and a decision frame skip of 4. We use successful capture rate (SR), mean task time cost (TC), mission-effective rule compliance (MRC), and heterogeneous coordination score (HCS) to provide a comprehensive assessment. 

Define the time-efficiency score as
\begin{equation}
    E_T=\min\left(1,\frac{T_0}{T}\right),
\end{equation}
where $T$ is the episode duration in seconds. $T_0=100\,\mathrm{s}$ is a predefined reference time, which is longer than the mean successful capture time of \method{} but shorter than the mean successful capture time of most algorithms and the duration of an ideal escape. 

To introduce mission-effective rule compliance (MRC), we first define the rule compliance rate (RCR) as 
\begin{equation}
    \mathrm{RCR}=\frac{1}{|\mathcal{N}|T}\sum_{t=1}^{T}\sum_{i\in\mathcal{N}}\mathbb{I}\{g_{i,t}\},
\end{equation}
where $\mathbb{I}(\cdot)$ denotes the indicator function. $g_{i,t}=1$ indicates that pursuer $i$ satisfies all navigation rules at time step $t$, whereas $g_{i,t}=0$ indicates at least one rule violation, including intrusion to shore, anchorage or buffer, and wrong-direction motion when it is inside an inbound or outbound lane. 

Then, MRC is computed as 
\begin{equation}
    \mathrm{MRC}=\mathrm{SR}\cdot \mathrm{RCR}\cdot E_T,
\end{equation}
MRC is more suitable for the constrained port pursuit task than using SR or RCR alone because the mission requires simultaneous capture success, navigation-rule compliance, and time efficiency. A policy may obtain a high RCR by behaving conservatively or avoiding active interception, but such behavior is ineffective if the evader escapes.

To measure the coordination performance of the heterogeneous system, we define HCS as
\begin{equation}
\mathrm{HCS}=0.30\mathrm{SR}+0.25E_T+0.25S_v+0.20I_c,
\end{equation}
where $S_v$ is the fraction of simulation steps in which at least one scout can sense the evader, and $I_c$ is the ratio of positive distance-closing contribution provided by the interceptors. All terms are normalized to $[0,1]$. HCS is introduced because task success alone cannot fully reflect whether the heterogeneous team performs coordinated role allocation. In the considered pursuit task, scouts are expected to maintain target visibility and update the shared evader belief, while interceptors are expected to reduce the evader distance and complete capture.

\begin{table}[t] \centering \caption{Evaluation Metrics over 100 Episodes} \label{tab:baseline} \resizebox{\linewidth}{!}{
\begin{tabular}{lccccc} \toprule 
\textbf{Algorithm} & \textbf{SR$\uparrow$ (\%)} & \textbf{95\% CI SR$\uparrow$ (\%)} & \textbf{TC$\downarrow$ (s)} & \textbf{MRC$\uparrow$} & \textbf{HCS$\uparrow$} \\
\midrule 
Expert Rules & 35.0 & [26.4, 44.7] & 164.75 & 0.1387 & 0.5348 \\ 
MADDPG & 2.0 & [0.6, 7.0] & 184.65 & 0.0090 & 0.3041 \\ 
OGR-MADDPG & 5.0 & [2.2, 11.2] & 192.38 & 0.0175 & 0.4459 \\ 
MATD3 & 1.0 & [0.2, 5.4] & 188.57 & 0.0034 & 0.3669 \\ 
OGR-MATD3 & 59.0 & [49.2, 68.1] & 142.37 & 0.2862 & 0.6163 \\ 
MAPPO & 16.0 & [10.1, 24.4] & 172.15 & 0.0557 & 0.3954\\
OGR-MAPPO & 72.0 & [62.5, 79.9] & \textbf{109.69} & \textbf{0.5114} & 0.6792\\
MASAC & 14.0 & [8.5, 22.1] & 182.92 & 0.0437 & 0.4244 \\ 
\textbf{OGR-MASAC} & \textbf{75.0} & \textbf{[65.7, 82.5]} & 121.94 & 0.4283 & \textbf{0.6802} 
\\ \bottomrule 
\end{tabular}
} 
\end{table}

\subsection{Baseline Comparison}
We evaluate \method{} by instantiating it with different MARL backbones, including MADDPG, MATD3, MAPPO, and MASAC. The resulting variants are denoted as OGR-MADDPG, OGR-MATD3, OGR-MAPPO and OGR-MASAC, respectively. These variants share the same option-guidance module, shared evader belief, action-blending mechanism, and reward design, while differing only in the residual MARL learner. Therefore, the comparison evaluates both the general applicability of \method{} and the influence of the selected residual learner.

The results in Table~\ref{tab:baseline} show that \method{} provides a broadly applicable option-guided residual learning interface for different MARL algorithms. The pure expert-rule controller reaches a success rate of 35.0\%, showing that rule-based option guidance alone is useful but insufficient against a fast and replanning evader. Pure MARL baselines perform poorly because direct continuous-control learning from scratch does not reliably discover the required cooperative behaviors under port-waterway constraints. Compared with their corresponding pure MARL baselines, the OGR variants generally improve SR, MRC, and HCS, indicating that the option layer supplies useful structural priors for constrained port pursuit. In particular, OGR-MATD3 improves the SR from 1.0\% to 59.0\%, and OGR-MASAC improves the SR from 14.0\% to 75.0\%. These results suggest that option guidance substantially reduces exploration difficulty and helps the learners discover harbor-exit, scout-tracking, and interceptor-pincer behaviors. Among the tested instantiations, OGR-MASAC achieves the best overall performance, with the highest SR, the highest HCS, and promising TC and MRC. This demonstrates that MASAC is the most effective residual learner within the proposed \method{} framework in the considered task.

\begin{figure}[t]
\centering
\includegraphics[width=\linewidth]{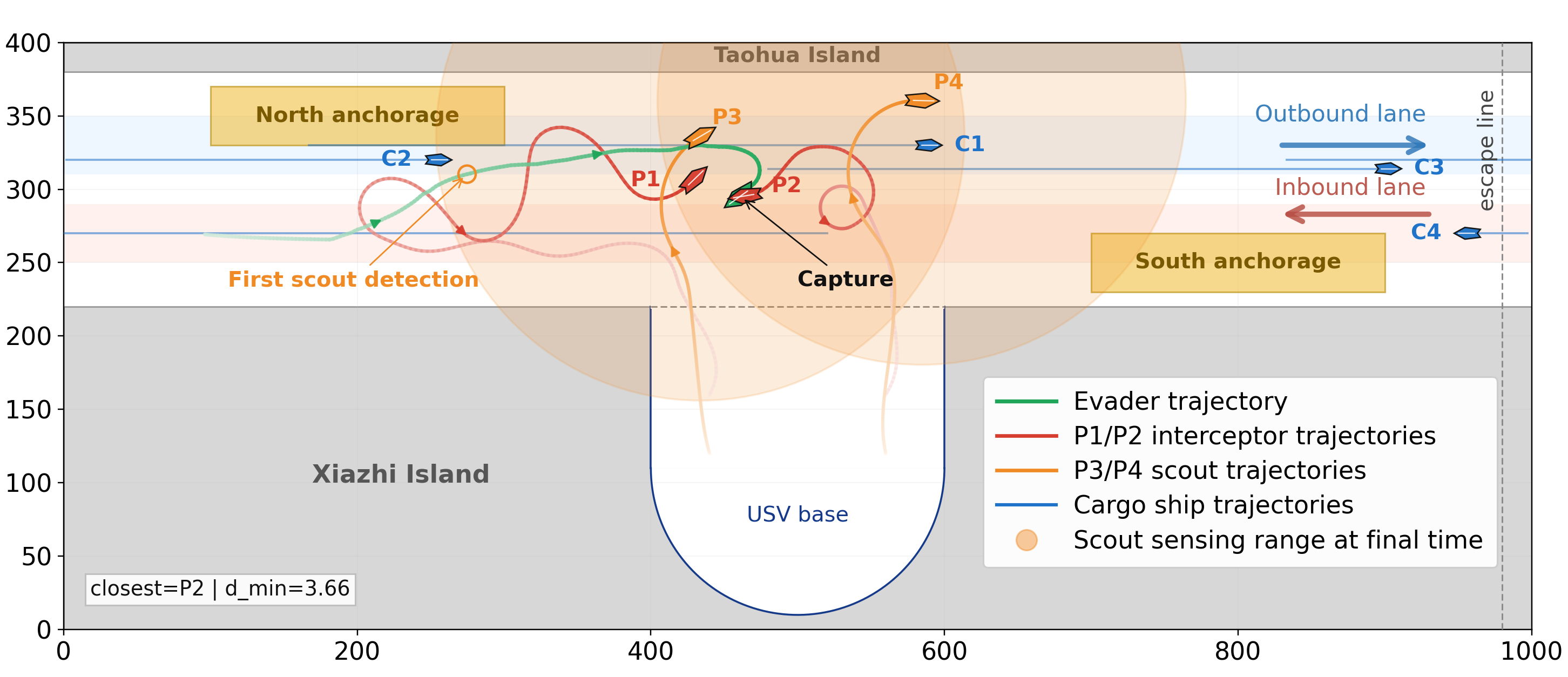}
\caption{A successful capture in the abstract Xiazhimen port waterways.}
\label{fig:abstract_success}
\end{figure}

Fig.~\ref{fig:abstract_success} presents a successful capture in the abstract Xiazhimen port waterways. The option-guided design encourages scouts to maintain evader visibility and interceptors to contribute to distance closing and final capture. We find that the remaining failures are mainly caused by inefficient terminal close-in interception rather than complete loss of the evader. Successful episodes finish with a mean closest interceptor--evader distance of 4.17, while failed episodes have a mean closest distance of 8.19. Moreover, scouts observe the evader for a larger fraction of time in failed episodes than in successful episodes, 0.738 versus 0.512, suggesting that the primary bottleneck is not target visibility but the final interception efficiency of the interceptors.

\subsection{Reward Ablation Study}
Table~\ref{tab:ablation} presents the reward ablation results. All algorithms retain the same option layer and architecture. Removing the rule reward or role reward reduces the performance, and removing both performs worst, confirming that rule and role rewards are effective.

\begin{figure}[t]
\centering
\includegraphics[width=\linewidth]{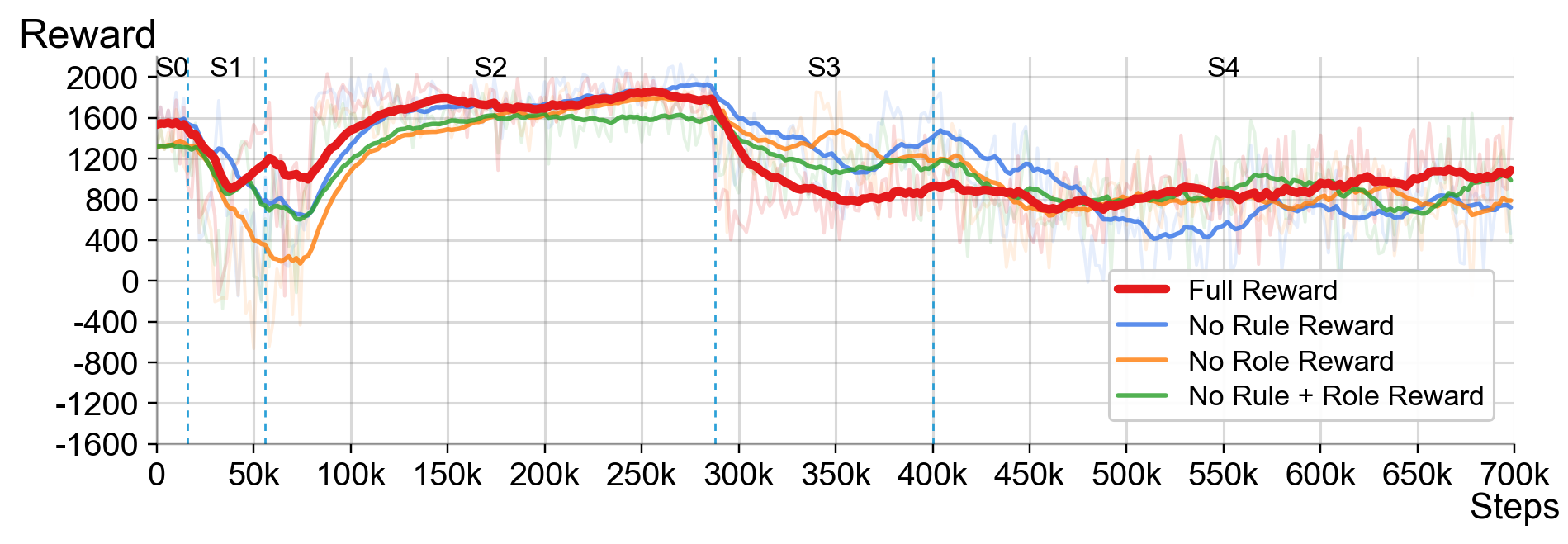}
\caption{Training reward curves for the reward-ablation variants. The full reward maintains the most stable improvement by 700k training steps.}
\label{fig:ablation}
\end{figure}

Fig.~\ref{fig:ablation} shows the reward curves of the ablation experiment during training. The reward degradations in the initial phase reflect the disruption caused to the prior policy by exploratory residual actions, while the drops around 300k steps are related to harder curriculum learning settings. Overall, the OGR-MASAC algorithm incorporating the full reward function maintains good training quality and successfully converges.
\begin{table}[t]
\caption{Reward Ablation Results over 100 Episodes}
\begin{center}
\resizebox{\linewidth}{!}{
\begin{tabular}{cccccc}
\toprule
\textbf{Algorithm} & \textbf{SR$\uparrow$ (\%)} & \textbf{95\% CI SR$\uparrow$ (\%)} & \textbf{TC$\downarrow$ (s)} & \textbf{MRC$\uparrow$} & \textbf{HCS$\uparrow$} \\
\midrule
\textbf{OGR-MASAC} & \textbf{75.0} & \textbf{[65.7, 82.5]} & \textbf{121.94} & \textbf{0.4283} & \textbf{0.6802}\\
w/o rule & 31.0 & [22.8, 40.6] & 164.26 & 0.1299 & 0.5214 \\
w/o role & 51.0 & [41.3, 60.6] & 144.32 & 0.2403 & 0.5953 \\
w/o rule+role & 28.0 & [20.1, 37.5] & 167.60 & 0.1040 & 0.5032 \\
\bottomrule
\end{tabular}
}
\label{tab:ablation}
\end{center}
\end{table}

\subsection{Zero-Shot Real-Map Transfer with AIS-Informed Traffic}

\begin{figure}[t]
\centering
\includegraphics[width=0.7\linewidth]{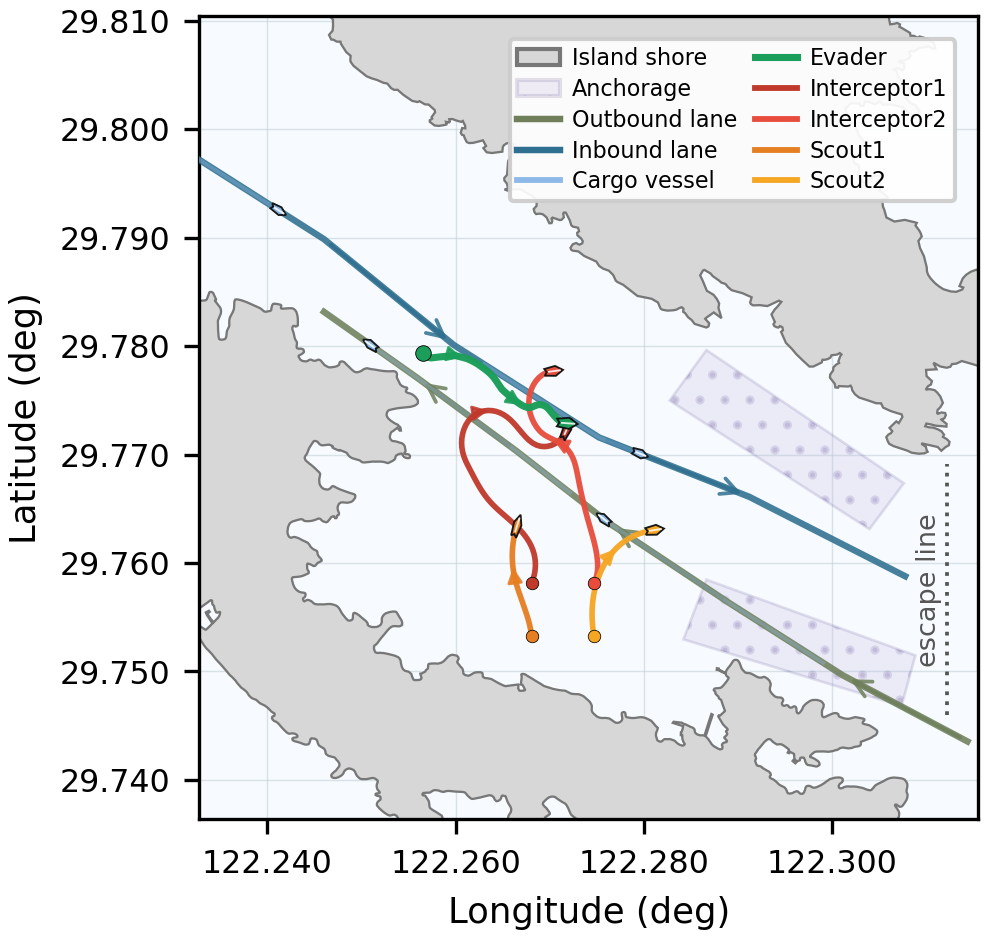}
\caption{A successful capture on the AIS-informed real map of Xiazhimen.}
\label{fig:realmap_capture}
\end{figure}

The zero-shot real-map transfer experiment uses a map of Xiazhimen in Zhejiang Province of China, where the \method{} algorithms are evaluated without retraining. The shores and anchorage areas are drawn in QGIS, while the traffic data is derived from AIS data \cite{b21, b24}. We consider four cargo vessels from AIS-C1 to AIS-C4 as dynamic obstacles. A successful capture is shown in Fig.~\ref{fig:realmap_capture}.

We evaluate OGR-MASAC over 100 test episodes on 15 fixed seeds without retraining. OGR-MASAC achieves 66.67\% of SR with $[41.7, 84.8]\%$ of CI. Successful captures finish in $31.65\pm 11.68\,\mathrm{s}$, while escaped episodes end in $126.30\pm 2.71\,\mathrm{s}$. Although OGR-MASAC achieves promising results, the SR in realistic maps is lower compared to the simulation experiments on the abstract map. This performance gap during the zero-shot real-map transfer is attributed to factors such as curved coastlines, irregular port layouts, irregularly moving cargo vessels, and initial position offsets.

\section{Conclusion}
This paper proposes \method{}, an algorithm-agnostic option-guided residual MARL framework for heterogeneous USV cooperative pursuit in constrained port waterways. Experiments demonstrate that it can be instantiated with different continuous-control MARL algorithms. In the abstract Xiazhimen scenario, the MASAC-based instantiation, namely OGR-MASAC, achieves a 75.0\% capture rate and obtains promising mission time cost, mission-effective rule compliance, and the highest heterogeneous coordination score among the tested methods. The QGIS/AIS-informed zero-shot real-map transfer experiment further demonstrates the potential generalization capability of the proposed framework without retraining. Future directions include improving sample efficiency using world models \cite{b26} and conducting further validation in more realistic or real-world maritime environments.

\end{document}